\documentclass{article}

 \usepackage[preprint]{neurips_2026}

\usepackage[utf8]{inputenc} 
\usepackage[T1]{fontenc}    
\usepackage{hyperref}       
\usepackage{url}            
\usepackage{booktabs}       
\usepackage{amsfonts}       
\usepackage{nicefrac}       
\usepackage{microtype}      
\usepackage[table,HTML]{xcolor}      

\usepackage{graphicx}
\usepackage{amsmath}
\usepackage{amssymb}
\usepackage{multirow}
\usepackage{subcaption}
\usepackage{algorithm}
\usepackage{algorithmicx}
\usepackage{algpseudocode}
\usepackage{caption}
\usepackage{stfloats}
\usepackage{float}
\usepackage{inconsolata}  
\usepackage{microtype}
\usepackage{enumitem}

\title{SnapMLA: Efficient Long-Context MLA Decoding via Hardware-Aware FP8 Quantized Pipelining}

\author{
  Yifan Zhang$^{1}$$^\dagger$\enskip\enskip
  Zunhai Su$^{1,2}$$^\dagger$\enskip\enskip
  Shuhao Hu$^{1}$$^\dagger$\thanks{Work done at Meituan.$^\dagger$ Equal Contribution. }\enskip\enskip
  Rui Yang$^{1}$\enskip\enskip\\
  \vspace{2mm}
  \textbf{Wei Wu$^{1}$\enskip\enskip
  Yulei Qian$^{1}$\enskip\enskip
  Yuchen Xie$^{1}$\enskip\enskip
  Xunliang Cai$^{1}$}\\
  $^{1}$Meituan LongCat Team \enspace
  $^{2}$Tsinghua University \enspace
}

\begin{document}

\maketitle

\vspace{-5mm}
\begin{abstract}
\vspace{-3mm}
While FP8 attention has shown substantial promise in innovations like FlashAttention-3, its integration into the decoding phase of the DeepSeek Multi-head Latent Attention (MLA) architecture presents notable challenges. 
These challenges include numerical heterogeneity arising from the decoupling of positional embeddings, misalignment of quantization scales in FP8 PV GEMM, and the need for optimized system-level support.
In this paper, we introduce \textit{SnapMLA}, an FP8 MLA decoding framework optimized to improve long-context efficiency through the following hardware-aware algorithm-kernel co-optimization techniques:
\textit{(i) RoPE-Aware Per-Token KV Quantization:} Motivated by our analysis of the heterogeneous quantization sensitivity inherent to the MLA KV cache, this approach preserves the RoPE part in high precision. 
Furthermore, per-token granularity is employed to align with the autoregressive decoding process and maintain quantization accuracy.   
\textit{(ii) Quantized PV Computation Pipeline Reconstruction:} Addresses the misalignment of quantization scales in FP8 PV computation caused by the shared KV structure of the MLA.  
\textit{(iii) End-to-End Dataflow Optimization:} Establishes an efficient data read-and-write workflow using specialized kernels, ensuring streamlined data flow and improved performance.
Extensive experiments on state-of-the-art MLA LLMs show that SnapMLA achieves up to \textbf{\textit{a 1.91× improvement in throughput on long-output decoding workloads while maintaining near-parity benchmark quality compared with the BF16 baseline}} on the evaluated reasoning and code-generation benchmarks.
Code is available at \href{https://github.com/meituan-longcat/SGLang-FluentLLM}{\texttt{https://github.com/meituan-longcat/SGLang-FluentLLM}}.
\end{abstract}


\begin{figure}[h]
    \centering

    \includegraphics[width=1\linewidth]{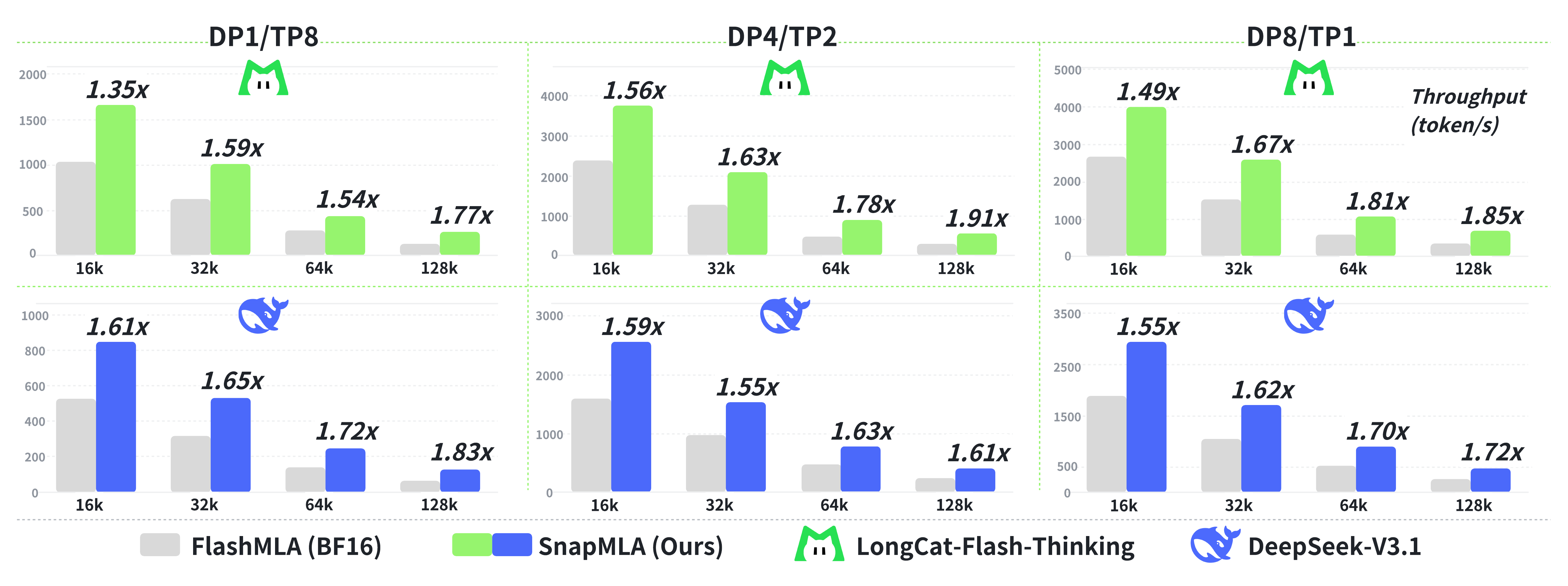}
    \caption{End-to-end decoding throughput comparison.
    We evaluate the generation throughput of SnapMLA on \textit{LongCat-Flash-Thinking} and \textit{DeepSeek-V3.1} across various parallelization configurations and context lengths.
    For additional details please refer to Section \ref{sec:system_performance}.}
    \vspace{-3mm}
    \label{fig:throughput}
\end{figure}
\section{Introduction}
Large language models (LLMs) have revolutionized the field of artificial intelligence\cite{annepaka2025large,liu2024deepseekv2,liu2025deepseekv3_2,liu2024deepseekv3,team2025longcat1,team2025longcat2, team2026longcat,team2025longcat3}, with their true potential increasingly dependent on the ability to efficiently process long-context data \cite{zhang2025efficient,hooper2024kvquant,xiong2025parallelcomp,xiong2025dope}. 
However, this advancement places significant strain on the underlying serving systems. 
As sequence length grows, the Key-Value (KV) cache, which stores intermediate attention states, expands linearly, rapidly consuming GPU memory and limiting both system throughput and the maximum supportable batch size \cite{shi2024keep,li2024survey}. 
Addressing this memory bottleneck has become critical for enabling efficient long-context inference \cite{hooper2024kvquant, su2025rotatekv,liu2024kivi,ye2025fit}.
To mitigate this issue, DeepSeek introduces the Multi-head Latent Attention (MLA) architecture \cite{liu2024deepseekv2,liu2024deepseekv3,liu2025deepseekv3_2,team2026longcat}, which reduces KV cache storage requirements through low-rank KV joint compression. 
However, in high-concurrency inference scenarios, architectural compression alone is insufficient. 
To further enhance throughput and support larger batch sizes, quantizing the KV cache into low-bit formats has emerged as a crucial strategy \cite{shi2024keep,li2024survey,liu2025deepseekv3_2, shah2024flashattention3,flashmla2025}.

With its exceptional representational capacity and impressive efficiency in both memory and computation, the FP8 data format for quantization has gained significant attention and widespread adoption \cite{micikevicius2022fp8,kuzmin2022fp8,van2023fp8,shah2024flashattention3,liu2024deepseekv3}. 
FP8 Tensor Cores, introduced with NVIDIA's Hopper architecture \cite{nvidia_h100_tensor_c}, are specifically designed to accelerate machine learning workloads using 8-bit floating point precision. 
However, fully harnessing the potential of FP8 during the decoding phase of MLA LLMs presents several distinct challenges:

\textbf{\textit{(i) FP8 KV Cache Quantization:}} 
MLA introduces architecture-induced numerical heterogeneity by decoupling KV vectors into a compressed content component and a precision-sensitive Rotational Position Embedding (RoPE) component. The application of uniform quantization does not effectively address this disparity.
Furthermore, when quantization is applied across tokens (e.g., at per-channel or per-block granularity), the autoregressive decoding prevents the immediate quantization of newly generated tokens. 
This entails complex buffer management, which undermines efficiency.

\textbf{\textit{(ii) FP8 MLA Computation:}} 
In MLA, a shared term is used to store and project both KVs.
This results in a misalignment of the quantization scale of \( V \) during the FP8 PV GEMM on Hopper Tensor Cores due to the \textbf{\textit{k-major layout constraint}} \cite{shah2024flashattention3}, rendering the standard post-GEMM dequantization process unfeasible.

\textbf{\textit{(iii) System-Level Support for FP8 MLA Decoding:}}
Effective FP8 MLA decoding depends on essential system-level optimizations. 
Minimizing memory access overhead and reducing kernel launch latencies are crucial for fully unlocking the potential of FP8 decoding, ensuring the effective realization of algorithmic advancements and other optimizations.

To address these challenges, we introduce \textit{SnapMLA}, an FP8 hardware-aware algorithm-kernel co-optimization framework designed to optimize long-context MLA decoding efficiency.
This framework tackles the identified challenges through three key components: \textit{RoPE-Aware Per-Token Quantization} (Section \ref{sec:method_quant}), \textit{Quantized PV Computation Pipeline Reconstruction} (Section \ref{sec:scale_pipeline}), and \textit{End-to-End Dataflow Optimization} (Section \ref{sec:method_fused}).
Our main contributions are summarized as follows:
\begin{itemize}
\item We introduce\textbf{\textit{ the first open-source FP8 decoding framework tailored to MLA decoding}}. 
By leveraging hardware-aware algorithm-kernel co-optimization, our framework effectively overcomes the challenges associated with MLA KV cache quantization, quantized MLA computation, and system-level support for FP8 MLA decoding.
\item Building on a comprehensive analysis of MLA KV cache numerical and quantization errors, we empirically identify, to the best of our knowledge for the first time, the divergent quantization sensitivity between the latent content and decoupled RoPE components in the MLA KV cache. 
We propose \textbf{\textit{RoPE-Aware Per-Token Quantization}}, along with supporting system optimizations, to ensure the preservation of the critical RoPE part from both algorithmic and system-level perspectives.
\item We develop custom CUDA kernels for SnapMLA, and extensive experiments on state-of-the-art MLA LLMs demonstrate that our approach achieves up to a \textbf{\textit{1.91× improvement in throughput on long-output decoding workloads (Figure \ref{fig:throughput}) while maintaining near-parity benchmark quality compared with the BF16 baseline (Table \ref{tab:benchmarks})}} on the evaluated reasoning and code-generation benchmarks.
\end{itemize}

\section{Preliminaries of Multi-Head Latent Attention}
\label{sec:mla_background}
Multi-Head Latent Attention (MLA), first introduced in DeepSeek-V2 \cite{liu2024deepseekv2}, has since been integrated into several state-of-the-art LLMs, including DeepSeek \cite{liu2024deepseekv3,liu2025deepseekv3_2,guo2025deepseek} and the LongCat-Flash series \cite{team2025longcat1,team2025longcat2,team2025longcat3,team2026longcat,liu2026scaling}.
The key innovation of MLA lies in the utilization of low-rank KV joint compression, which projects both KVs into a shared, low-dimensional latent space. 
In the following, we explore the core components of MLA.
\paragraph{Low-Rank Compression and Decoupled RoPE}  
Let $ {h}_t \in \mathbb{R}^{d}$ denote the input hidden state at time step $t$. 
MLA begins by compressing the joint KV  representation into a latent vector $ {c}_{KV} \in \mathbb{R}^{d_{c}}$ using a down-projection matrix $W^{DKV} \in \mathbb{R}^{d_{c} \times d}$:
\begin{equation}
     {c}_{KV} = W^{DKV}  {h}_t
\end{equation}
where $d_c$ is the compression dimension, with $d_c \ll d_h n_h$.
To resolve the incompatibility between low-rank compression and position-sensitive embeddings, MLA employs a \textit{Decoupled RoPE} strategy. 
The \( K \) tensor for each attention head \( i \), is partitioned into two components:
a compressed content part $ {k}_{i}^C$ and a RoPE part $ {k}^R$:
\begin{equation}
     {k}_{i}^C = W^{UK}  {c}_{KV}, \quad  {k}^R = \text{RoPE}(W^{KR}  {h}_t)
\end{equation}
Here, \( W^{UK} \) is the up-projection matrix for the \( K \), and \( W^{KR} \) is the matrix for the RoPE term. 
\(  {k}^R \) carries the RoPE, which is shared across all heads.
The final \( K \) used for attention is the concatenation of $ {k}_{i}^C$ and $ {k}^R$:
\begin{equation}
     {k}_i = [ {k}_{i}^C;  {k}^R]
\end{equation}
The \( V \) vector is derived solely from the latent vector:
\begin{equation}
     {v}_{i}^C = W^{UV}  {c}_{KV}
\end{equation}

\paragraph{Inference-Optimized Absorbed Mode}
The matrix absorption technique is a crucial process in MLA that ensures inference efficiency.
During the decoding phase, the up-projection matrices for \( K \) (\( W^{UK} \)) and \( V \) (\( W^{UV} \)) are integrated into the \( Q \) projection (\( W^Q \)) and the output projection (\( W^O \)), respectively.
This optimization eliminates the need for explicit reconstruction of the compressed content components $ {k}^C$ and $ {v}^C$. 
As a result, the attention score calculation for the $i$-th head between the current Query $ {q}_{t,i}$ and a past token $j$ is reformulated as:
\begin{equation}
q_{t,i}^T k_{j,i} = \underbrace{( {q}_{t,i}^C W^{UK})^T  {c}_{KV, j}}_{\text{Content Term}} + \underbrace{( {q}_{t,i}^R)^T  {k}^R_j}_{\text{RoPE Term}}
\label{eq:absorbed_attention}
\end{equation}
\paragraph{Challenges of MLA Quantization}
The distinctive attention structure of MLA presents two major challenges in achieving efficient quantization. 
First, the MLA architecture splits the KV cache for each token into two distinct components, each serving a specific function:
\begin{itemize}
    \item \textbf{\textit{Compressed Latent Vector \( {c}_{KV} \in \mathbb{R}^{d_c} \):}} A dense, low-dimensional vector containing contextual content information.
    \item \textbf{\textit{Decoupled RoPE \(  {k}^R \in \mathbb{R}^{d_h^R} \):}} A position-sensitive vector that requires high precision to effectively preserve positional information.
\end{itemize}
The functional heterogeneity between these components introduces numerical heterogeneity, leading to differences in quantization sensitivity—an issue not extensively explored in prior work.

Second, executing the PV GEMM under FP8 precision on Hopper Tensor Cores requires adherence to a strict \textbf{\textit{k-major layout constraint}} \cite{shah2024flashattention3}. 
However, MLA introduces a distinct challenge: \( V \) inherits per-token quantization scales from the latent cache (\( C_{KV} \)), which are aligned along the \textit{reduction dimension}. This alignment disrupts the conventional post-GEMM dequantization process.

Along with the preliminaries of MLA, the preliminaries of low-bit quantization are presented in Appendix \ref{appendix:low bit quant}.

\section{Methodology}
\label{sec:method}
\begin{figure}[t]
    \centering
    \vspace{3mm}
    \includegraphics[width=1\linewidth]{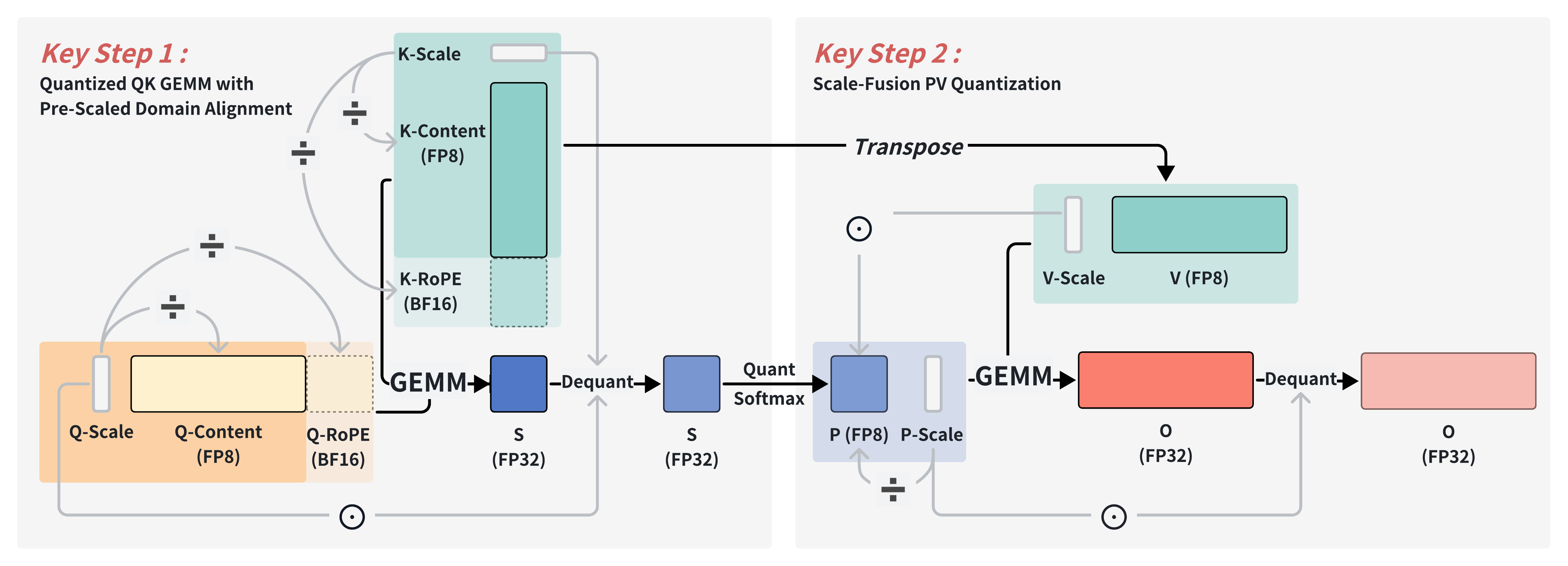}
    \caption{Overview of the scale fusion pipeline in SnapMLA. Note that $K$ and $V$ content share the latent cache (\( \mathbf{c}_{KV} \)).
    \textbf{\textit{Key Step 1}} illustrates \textit{RoPE-Aware Per-Token KV Quantization}, where the BF16 RoPE part of the QK is pre-scaled with the quantization scale of the content part, thereby unifying the numerical domains of the quantized content part and unquantized RoPE part.
    \textbf{\textit{Key Step 2}} demonstrates \textit{Quantized PV Computation Pipeline Reconstruction}, where the scale of \( V \) is pre-fused into the \( P \), circumventing quantization dimension mismatch.}
    \label{fig:fuse_scale_attn}
\end{figure}

In this section, we present \textit{SnapMLA}, a hardware-aware algorithm-kernel co-optimization framework designed to optimize long-context MLA decoding. 
We begin by introducing \textit{RoPE-Aware Per-Token KV Quantization} in Section \ref{sec:method_quant}. 
In Section \ref{sec:scale_pipeline}, we explore \textit{Quantized PV Computation Pipeline Reconstruction}, which resolves the misalignment of quantization scales.
In Section \ref{sec:method_fused}, we introduce \textit{End-to-End Dataflow Optimization}, where we establish a streamlined data read-and-write workflow.
Figure \ref{fig:fuse_scale_attn} provides an overview of the critical scale fusion pipeline in SnapMLA.
The algorithm for the SnapMLA decoding process is presented in Algorithm~\ref{alg:flashmla_fp8_dual_wg}.

\subsection{RoPE-Aware Per-Token KV Quantization}
\label{sec:method_quant}
Building upon a comprehensive analysis of numerical values and quantization errors in the MLA KV cache, we introduce \textit{RoPE-Aware Per-Token Quantization} in this section, supported by targeted system-level optimizations to enhance both its efficiency and performance.

\subsubsection{MLA KV Quantization: An Algorithmic Perspective}
\begin{figure}[t]
    \centering
    \begin{subfigure}[t]{0.48\linewidth}
        \centering
        \includegraphics[width=\linewidth]{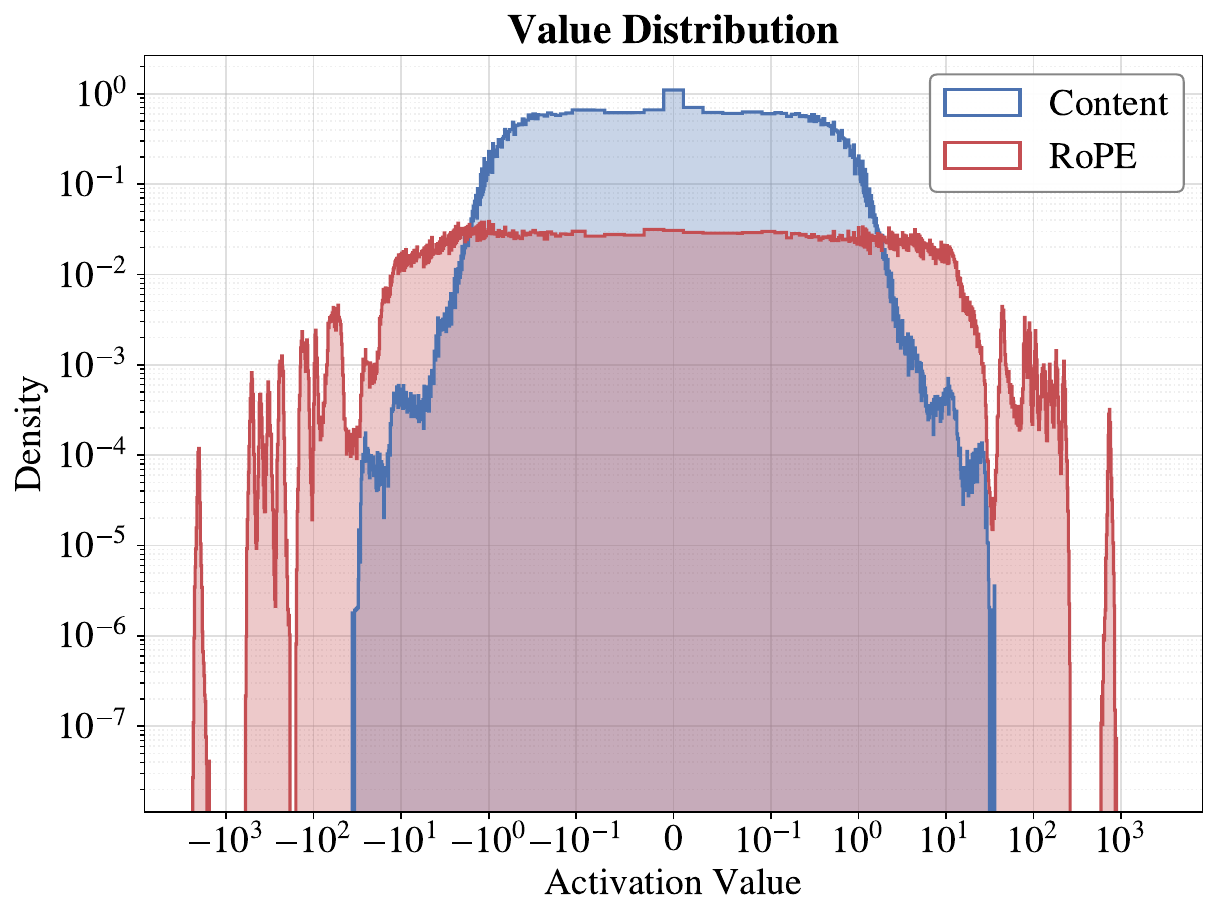}
        \caption{Numerical values analysis.}
        \label{fig:rope_distribution_a}
    \end{subfigure}
    \hfill 
    \begin{subfigure}[t]{0.48\linewidth}
        \centering
        \includegraphics[width=\linewidth]{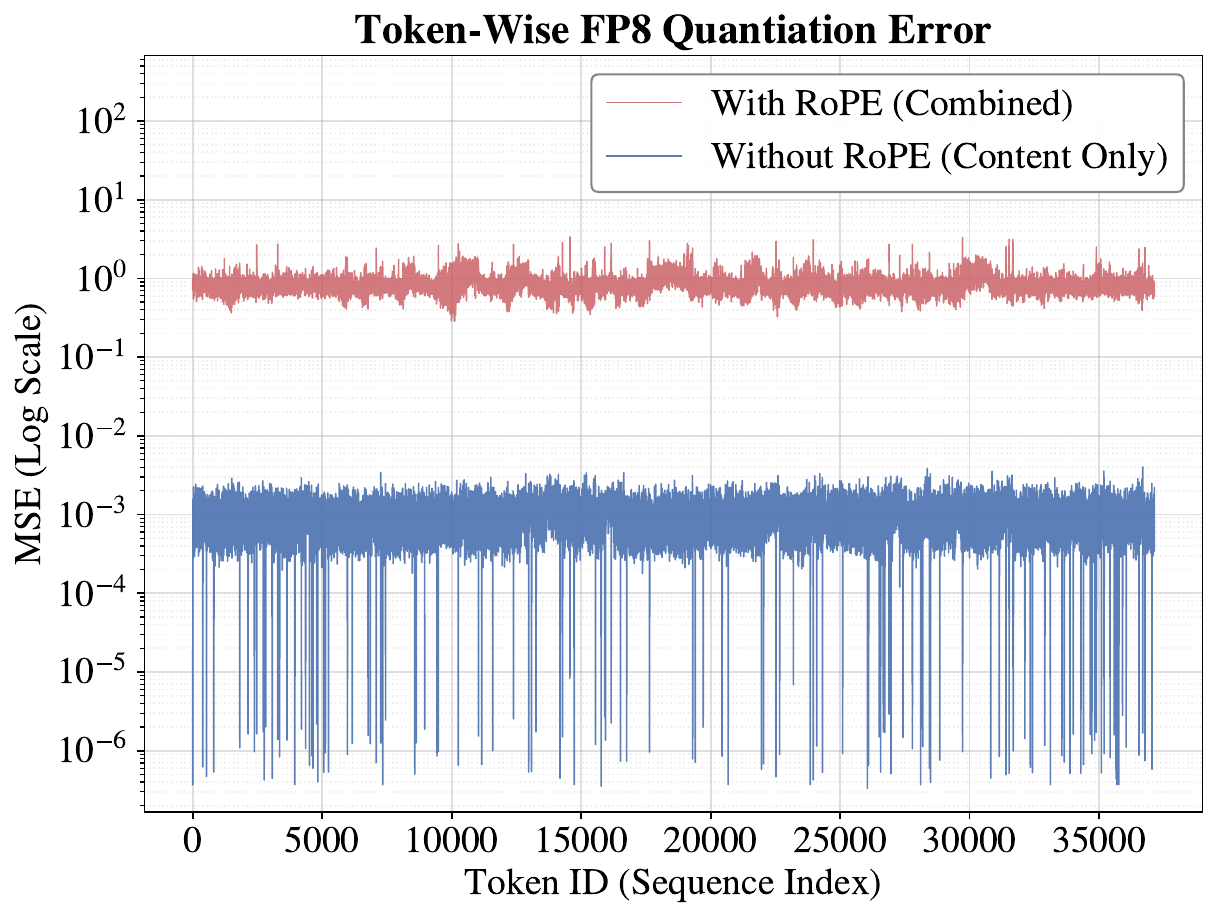}
        \caption{Quantization error analysis..}
        \label{fig:rope_distribution_b}
    \end{subfigure}

    \caption{Analysis of the numerical value distribution and quantization error comparison for the content and RoPE components of MLA KV cache in \textit{LongCat-Flash-Thinking}.}
    \label{fig:rope_distribution}
\end{figure}

\paragraph{Heterogeneous Quantization Sensitivity}
The structure of the MLA KV cache represents a notable departure from traditional multi-head attention (MHA), as detailed in Section \ref{sec:mla_background}.
Effective and reliable quantization hinges on a deep understanding of the numerical distribution and a thorough re-evaluation of quantization sensitivity. 
To this end, we conduct an extensive analysis of the content and RoPE components of the MLA KV cache, based on \textit{LongCat-Flash-Thinking} \cite{team2025longcat2}.
Our empirical findings uncover a pronounced divergence in the distributional characteristics of these two parts. 
As shown in Figure~\ref{fig:rope_distribution_a}, the RoPE component spans a significantly wider dynamic range (reaching $\pm 10^3$), exhibiting distinct outlier tails, while the content component is tightly concentrated around zero (within $\pm 10^1$), highlighting a stark contrast in their statistical behavior.

This scale mismatch renders uniform quantization strategies ineffective. 
Specifically, FP8 quantization leads to an order-of-magnitude increase in Mean Squared Error (MSE) for the RoPE component (as shown in Figure~\ref{fig:rope_distribution_b}), highlighting their heightened sensitivity to quantization.
To address these issues, we propose a \textit{RoPE-aware} strategy that applies FP8 quantization exclusively to the content components, while retaining the RoPE components in BF16 to preserve positional robustness.
Although this quantization scheme is straightforward, it is firmly grounded in analysis and plays a pivotal role in preserving accuracy while optimizing KV cache compression rates.
 
\paragraph{Decoding-Centric Quantization Granularity for Accuracy Preservation}
While approaches like FA3 \cite{shah2024flashattention3} demonstrate effective FP8 quantization during the prefill stage through block-wise quantization, this method encounters limitations in autoregressive decoding. 
In the decoding stage, tokens are generated sequentially, leading to "page tail" tokens within incomplete KV quantization blocks. 
This requires complex buffer management, which renders block-level quantization inefficient. 
To address this, we adopt per-token quantization granularity, offering two primary advantages:
\begin{itemize}
    \item \textbf{\textit{Instant quantization}}, which enables immediate quantization of newly generated KV tokens and eliminates the need for "tail buffer" management.
    \item \textbf{\textit{Framework compatibility}}, which allows processing all tokens with unified logic, thereby simplifying integration into inference frameworks (e.g., vLLM, SGLang).
\end{itemize}

Building on the comprehensive algorithmic insights presented above, we adopt \textit{RoPE-Aware Per-Token KV Quantization} as the foundation of our core quantization strategy.
\subsubsection{Hardware-Optimized RoPE-Aware Per-Token Quantization}
\label{sec:method_kernel}

\paragraph{Eliminating Mixed-Precision Accumulation in Quantized QK GEMM}
While the algorithmic details of \textit{RoPE-Aware Per-Token KV Quantization} were discussed in the previous section, its efficient hardware implementation introduces the challenge of mixed-precision accumulation in quantized QK GEMM. 
The basic MLA implementation (i.e., FlashMLA \cite{flashmla2025}) leverages instruction-level parallelism by partitioning the QK GEMM reduction dimension into nine thread groups (each spanning 64 dimensions) and employing a permuted execution schedule. 
The RoPE-aware quantization method introduces an inefficient mixed-precision constraint: the first eight groups (content components) use FP8 and require dequantization, while the ninth group (RoPE component) retains BF16 precision.

A potential solution is mixed-precision accumulation, where FP8 blocks are first aggregated, followed by dequantization and the addition of RoPE. 
However, this approach introduces a strict synchronization barrier, preventing interleaved execution, which is essential for optimal pipelining. 
As a result, pipeline bubbles are introduced, leading to a reduction in the actual throughput gains.

\paragraph{Quantized QK GEMM with Pre-Scaled Domain Alignment}
Instead of modifying the kernel's accumulation logic, we strategically project the BF16 RoPE components into the FP8 quantization domain during the data preparation phase.
As shown in \textbf{\textit{Key Step 1}} of Figure~\ref{fig:fuse_scale_attn}, we pre-scale the RoPE terms by the inverse of the content quantization scale:
\begin{equation}
\mathbf{Q}^R = \frac{\mathbf{Q}^R}{\mathbf{S}^{Q_c}}, \quad \mathbf{K}^R = \frac{\mathbf{K}^R}{\mathbf{S}^{K_c}}.
\end{equation}
This alignment unifies the numerical domains, effectively “disguising” the high-precision RoPE segments as scaled quantized values. 
As a result, the kernel can process all groups uniformly, reusing the original highly optimized accumulation order without the need for intermediate dequantization or synchronization steps. 

\label{sec:tiling_transpose}

\subsection{Quantized PV Computation Pipeline Reconstruction}
\label{sec:scale_pipeline}

\subsubsection{Memory Layout Constraints and Quantization Mismatch}
The execution of the second GEMM (PV) under FP8 precision on Hopper Tensor Cores requires strict adherence to the \textbf{\textit{k-major layout constraint}} for \( V \) \cite{shah2024flashattention3}. 
This hardware constraint mandates that \( V \) must be contiguous along the sequence dimension. 
However, in the context of MLA, \( V \) inherits per-token quantization scales from the latent cache (\( \mathbf{c}_{KV} \)). 
These scales are aligned with the GEMM's \textit{reduction dimension}, resulting in a fundamental architectural conflict. 
Consequently, the conventional post-GEMM dequantization paradigm fails to work as intended.

\subsubsection{Scale-Fusion PV Quantization: A Hardware-Algorithm Co-Optimization}

To address the quantization scale mismatch, we propose a unified algorithmic design that seamlessly integrates three tightly coupled components, collectively enabling efficient and hardware-optimized quantized PV computation:

\textit{\textbf{(i) Scale Fusion:}  }
In the first step, we resolve the scale mismatch by leveraging the associative property of multiplication (see \textbf{\textit{Key Step 2}} in Figure~\ref{fig:fuse_scale_attn}). 
The quantization scale of \( V \), denoted \( S_V \), is fused into the attention probability matrix \( P \), resulting in the scaled matrix \( P' = P \odot S_V \).

\textit{\textbf{(ii) Block-Wise Dynamic P Quantization:}}  
Since incorporating \( S_V \) modifies the original data distribution and extends the dynamic range of \( P \), we apply block-wise dynamic quantization to \( P' \) to maintain robust quantization accuracy. 
The block size is set to the PV GEMM kernel's tiling parameter (\( BlockN = 64 \)), ensuring numerical precision across dynamic value ranges while seamlessly integrating with the kernel’s tiled execution pattern.

\textit{\textbf{(iii) Implicit Dequantization:}}  
Block-wise \( P \) quantization introduces scale variations across blocks. To preserve arithmetic accuracy during tiled accumulation without incurring explicit quantization scaling overhead, we integrate the quantization scaling operation into the online Softmax accumulation states. 
This integration allows the tiled accumulation to account for scale differences across blocks. 
As a result, the online accumulation realizes \textit{implicit dequantization}, ensuring correctly scaled outputs within a unified computational flow. 
Importantly, this fusion does not modify the QK logits or the resulting Softmax distribution; the scale handling is applied to the post-Softmax probability blocks and their accumulation states.
A more detailed description of the scale-aware online accumulation mechanism is provided in Appendix \ref{Online Scale Fusion in the Softmax Function}.

\subsubsection{Summary of the Quantized PV Computation Pipeline}
We present the reconstructed pipeline as four block-wise stages:

\begin{enumerate}[topsep=2pt, itemsep=2pt, parsep=2pt]
    \item \textbf{Online Softmax}: Process the restored QK logits block by block, updating the running maximum and normalization statistic for the global Softmax.
    \item \textbf{Scale Fusion}: Integrate the per-token \( V \) scale \( S_V \) into the current probability block \( P \) to obtain the fused probability block \( P' = P \odot S_V \).
    \item \textbf{Quantization}: Apply block-wise dynamic quantization to \( P' \), producing a quantized probability block and its scale.
    \item \textbf{Computation}: Perform tiled PV GEMM and update \( O \) with scale-aware accumulation, which implicitly accounts for the probability-block quantization scale.
\end{enumerate}

Together, these co-designed components seamlessly integrate scale management into the attention computation, eliminating traditional scale mismatches and dequantization overhead, while fully adhering to hardware execution constraints. 
Additional implementation details for the quantized PV computation can be found in Appendix \ref{Supplementary Implementation Details for Quantized PV Computation}.


\subsection{End-to-End Dataflow Optimization}

\label{sec:method_fused}

While Section~\ref{sec:scale_pipeline} addresses the computational challenges in quantized attention, this section focuses on mitigating critical \textit{memory and data movement bottlenecks} in the end-to-end pipeline. We propose a systematic optimization strategy consisting of three interconnected layers: (i) fused kernel design to eliminate intermediate memory traffic, (ii) memory subsystem optimization via cache-aware tiling, and (iii) zero-overhead data layout transformation to meet Tensor Core requirements.

\subsubsection{Layer 1: Fused Compute-Memory Operators}

\paragraph{Fused Token Preparation: Q-Quant and K-Append}
We replace the traditional sequential workflow—comprising statistic computation, quantization, and memory copying—with two specialized atomic kernels. 
The \textbf{\textit{Fused-Q-Quant}} kernel consolidates per-token scale calculation, mixed-precision conversion, and \textit{Scale Domain Alignment} (by injecting quantization scales into RoPE dimensions) into a single operation.
The \textbf{\textit{Fused-K-Append}} kernel extends this paradigm to KV cache management. In addition to performing quantization and alignment, it integrates PagedAttention-style non-contiguous writes directly into the computational stream. 
This enables the system to complete quantization, scale alignment, and cache updates in a single kernel launch, thereby eliminating intermediate memory buffers and reducing launch overhead.

\paragraph{Fused Fetch-and-Dequant for Chunk Processing}
For decoding phases that require high-precision reuse of cached data (e.g., chunk prefill, prefix caching), we introduce the \textbf{\textit{Fused-Fetch-Dequant}} operator. 
This kernel performs on-the-fly, register-level dequantization immediately after fetching data from the quantized KV cache. 
By combining the dequantization operation directly with the load instructions, we streamline the memory retrieval and precision conversion into a single step. 
This eliminates the need for the traditional two-step process—loading quantized data into shared memory followed by a separate dequantization kernel—thereby reducing global memory I/O and kernel scheduling latency.

\subsubsection{Layer 2: Memory Subsystem Optimization}

\paragraph{Cache-Aligned Tiling and Swizzling}
To fully leverage the benefits of FP8’s reduced bit-width, we increase the tiling size along the content dimension from 64 to 128. 
This strategic adjustment ensures that each atomic memory load operation is aligned with the 128-byte L2 cache lines and the Swizzle-128B shared memory (SMEM) layout in the Hopper architecture. 
As a result, this access pattern generates fully coalesced Tensor Memory Accelerator (TMA) descriptors, optimizing High Bandwidth Memory (HBM) utilization and eliminating shared memory bank conflicts.

\subsubsection{Layer 3: Zero-Overhead Data Layout Transformation}

To satisfy the strict \textbf{\textit{k-major layout constraint}} required by FP8 WGMMA instructions, we implement an in-kernel transformation strategy tailored to our quantized pipeline. 
This involves two synchronized, fine-grained operations:

\textit{\textbf{(i) V-Tile Transposition via Register File:}} V tensor tiles are transposed using a SMEM $\rightarrow$ RF $\rightarrow$ SMEM routing path, leveraging the Register File for low-latency data reordering.

\textit{\textbf{(ii) P-Accumulator Byte Permutation:}} The accumulator holding intermediate attention scores undergoes byte-level register permutations to match the WGMMA output layout.

Critically, we schedule these data movements to leverage Hopper's asynchronous execution capabilities. 
By overlapping the \(V\) transposition and \(P\) permutation with the computation intervals of the preceding QK GEMM, we effectively mask their latency. 
This fine-grained overlap, combined with the extended $64 \times 128$ tiling scheme, enables zero-overhead layout adaptation while maintaining full compatibility with the mixed-precision pipeline.

\section{Experiments}
\label{sec:experiments}
In this section, we present an extensive evaluation of SnapMLA, assessing its effectiveness from three crucial perspectives: 
(i) \textit{Benchmark Results} (Section \ref{sec:benchmark}), where we rigorously evaluate SnapMLA across a wide range of tasks, including general QA, mathematical reasoning, general reasoning, coding, and others, to ensure its robustness across diverse domains;
(ii) \textit{Numerical Accuracy} (Appendix \ref{appendix:numerical_accuracy}), focusing on the precision of SnapMLA to verify its ability to preserve data fidelity during quantization;
and (iii) \textit{Efficiency Analysis} (Section \ref{sec:system_performance}), where we evaluate the efficiency of SnapMLA.

\subsection{Experimental Setup}
\label{sec:setup}

\paragraph{Models and Baselines}
We evaluate state-of-the-art MLA LLMs from two different series: \textit{DeepSeek-V3.1} \cite{liu2024deepseekv3} and \textit{LongCat-Flash-Thinking} \cite{team2025longcat1}. 
\textit{DeepSeek-V3.1} is a mixture-of-experts (MoE) LLM with 671 billion total parameters, activating 37 billion parameters per token. 
\textit{LongCat-Flash} is a 560-billion-parameter MoE LLM optimized for computational efficiency. 
It dynamically allocates computational resources by activating between 18.6 billion and 31.3 billion parameters per token, depending on the \textit{Zero-Computation Expert} mechanism. 
Additionally, it enhances the computation-communication overlap through the \textit{Shortcut-connected MoE} design.

We use FlashMLA \cite{shah2024flashattention3} (the standard BF16 implementation) as our baseline, serving as the reference for both benchmark accuracy and efficiency performance.
Details of the benchmarks used in the experiments are provided in Appendix \ref{appendix_benchmark}.
FlashMLA is a specialized attention kernel optimized for the MLA proposed by DeepSeek \cite{liu2024deepseekv3}, specifically tailored for NVIDIA Hopper GPUs.
For the FP8 experiments, we adopt the E4M3 format for efficient KV cache and MLA computation. 

Because SnapMLA targets decoding attention rather than model weights, the benchmark scores in Table \ref{tab:benchmarks} evaluate whether the FP8 decoding pipeline preserves model quality on representative tasks.
Many of the evaluated reasoning and coding benchmarks produce long generations, exercising the decoding kernel over thousands to tens of thousands of generated tokens; Appendix \ref{appendix_benchmark} reports their average generated lengths to characterize these long-output decoding workloads.
Separately, Section \ref{sec:system_performance} sweeps runtime KV-cache lengths from 16k to 128k to evaluate throughput under controlled serving configurations.


\begin{table*}[t]
\centering
\caption{Benchmark evaluation results across various task domains.}

\label{tab:benchmarks}
\resizebox{1\textwidth}{!}{%
\begin{tabular}{@{}lcccc@{}}
\toprule
\multicolumn{1}{l|}{} & \multicolumn{2}{c}{DeepSeek-V3.1} & \multicolumn{2}{c}{LongCat-Flash-thinking} \\ \cmidrule(l){2-5} 
\multicolumn{1}{l|}{} & FlashMLA & \textbf{SnapMLA (Ours)} & FlashMLA & \textbf{SnapMLA (Ours)} \\
\multicolumn{1}{l|}{\multirow{-3}{*}{Benchmark}} & BF16 & \textbf{FP8} & BF16 & \textbf{FP8} \\ \midrule
\multicolumn{5}{c}{\cellcolor[HTML]{EFEFEF}\textit{General QA}} \\
\multicolumn{1}{l|}{MMLU-Pro$_{\text{(acc)}}$} & 84.41 & 84.43 & 82.60 & 82.73 \\
\multicolumn{1}{l|}{MMLU-Redux$_{\text{(acc)}}$} & 90.48 & 90.89 & 89.30 & 88.27 \\ \midrule
\multicolumn{5}{c}{\cellcolor[HTML]{EFEFEF}\textit{Alignment}} \\
\multicolumn{1}{l|}{IFEval$_{\text{(strict prompt)}}$} & 86.32 & 87.25 & 86.90 & 87.80 \\
\multicolumn{1}{l|}{Arena-Hard$_{\text{(hard prompt gemini)}}$} & 57.10 & 55.50 & 69.90 & 70.40 \\ \midrule
\multicolumn{5}{c}{\cellcolor[HTML]{EFEFEF}\textit{Mathematical Reasoning}} \\
\multicolumn{1}{l|}{MATH-500$_{\text{(Mean@1)}}$} & 98.80 & 98.20 & 99.20 & 98.80 \\
\multicolumn{1}{l|}{AIME-24$_{\text{(Mean@32)}}$} & 93.85 & 93.65 & 91.87 & 91.67 \\
\multicolumn{1}{l|}{AIME-25$_{\text{(Mean@32)}}$} & 87.92 & 85.42 & 89.58 & 88.44 \\
\multicolumn{1}{l|}{BeyondAIME$_{\text{(Mean@10)}}$} & 71.80 & 69.90 & 69.50 & 70.20 \\ \midrule
\multicolumn{5}{c}{\cellcolor[HTML]{EFEFEF}\textit{General Reasoning}} \\
\multicolumn{1}{l|}{GPQA-Diamond$_{\text{(Mean@16)}}$} & 84.15 & 82.57 & 81.50 & 80.24 \\
\multicolumn{1}{l|}{ZebraLogic$_{\text{(Mean@1)}}$} & 96.10 & 96.00 & 95.50 & 95.00 \\ \midrule
\multicolumn{5}{c}{\cellcolor[HTML]{EFEFEF}\textit{Coding}} \\
\multicolumn{1}{l|}{LCB (24.08-25.05)$_{\text{(Mean@4)}}$} & 73.46 & 72.74 & 79.40 & 79.74 \\ \bottomrule
\end{tabular}
}
\end{table*}



\paragraph{Infrastructure Details} 
All experiments are conducted on a high-performance node equipped with 8 \(\times\) NVIDIA Hopper architecture GPUs\footnote{Due to certain restrictions, the specific type of the Hopper series GPU used cannot be disclosed; however, the full codebase is publicly available at \href{https://github.com/meituan-longcat/SGLang-FluentLLM}{\texttt{https://github.com/meituan-longcat/SGLang-FluentLLM}}
to ensure full reproducibility of the reported results.}.

\subsection{Benchmark Results}
\label{sec:benchmark}
\paragraph{Main Results}
As shown in Table \ref{tab:benchmarks}, SnapMLA maintains near-parity benchmark quality relative to the BF16 FlashMLA baseline across various task domains, even with FP8 quantization. 
The results demonstrate that SnapMLA preserves accuracy across general QA, alignment, and reasoning tasks, highlighting the effectiveness of our sensitivity-aware mixed-precision strategy. 
In mathematical reasoning tasks, SnapMLA remains close to the BF16 baseline, while in coding tasks, it either matches or slightly surpasses FlashMLA. 
Overall, SnapMLA preserves benchmark quality across diverse tasks, validating its ability to maintain accuracy while reducing computational overhead.
Experimental results on kernel numerical precision are provided in Appendix \ref{appendix:numerical_accuracy}, demonstrating that our method maintains numerical accuracy comparable to the BF16 baseline.


\subsection{Efficiency Analysis}
\label{sec:system_performance}

\paragraph{End-to-End Throughput}
Figure \ref{fig:throughput} reports end-to-end decoding throughput under matched per-rank input shapes for BF16 FlashMLA and SnapMLA.
We assess generation throughput across different parallelism strategies, examining both data parallelism (DP) and tensor parallelism (TP) configurations (DP1/TP8, DP4/TP2, DP8/TP1), as well as varying context lengths ranging from 16k to 128k.

As shown in Figure \ref{fig:throughput}, SnapMLA consistently outperforms FlashMLA. 
The largest observed speedup reaches up to 1.91$\times$, demonstrating the execution efficiency of the FP8 MLA decoding pipeline under matched workload shapes.
Figure \ref{fig:kernel_configurations} further isolates kernel-level TFLOPS under fixed-batch synthetic configurations; no batch-size search is used in that sensitivity study.
Additional efficiency analyses, including kernel Roofline performance and input configuration sensitivity, can be found in Appendix \ref{appendix:roofline_analysis} and \ref{appendix:sensitivity_input}.
Appendix \ref{appendix:design_evidence} summarizes how the numerical, end-to-end, and kernel-level evaluations support the coupled design choices in SnapMLA, while also clarifying that they are not intended as a complete factorial ablation of every kernel subcomponent.


\section{Related Work}
\label{sec:related_work}
\paragraph{FP8 Attention and MLA Decoding}
FP8 attention kernels such as FlashAttention-3~\cite{shah2024flashattention3} demonstrate that Hopper FP8 Tensor Cores can substantially accelerate attention when the computation is organized around hardware layout constraints such as the k-major operand format. However, these designs target standard attention and do not directly address MLA's shared latent KV cache or its decoupled RoPE component, whose numerical sensitivity motivates the mixed-precision treatment in SnapMLA. FlashMLA~\cite{flashmla2025} provides a highly optimized BF16 MLA decoding kernel on Hopper GPUs, but directly extending it to FP8 introduces scale-layout mismatches and pipeline hazards under per-token quantization. SnapMLA builds on these lines of work by co-designing RoPE-aware KV-cache quantization, scale-aware FP8 PV computation, and fused data movement for MLA decoding. Additional related-work discussion is provided in Appendix~\ref{appendix:additional_related_work}.

\section{Conclusion}
\label{sec:conclusion}
In this paper, we introduce SnapMLA, an FP8 decoding framework specifically designed to optimize long-context inference for MLA-based LLMs. 
By leveraging hardware-aware algorithm-kernel co-optimization, our framework effectively addresses the challenges associated with MLA KV cache quantization, quantized MLA computation, and system-level support for FP8 MLA decoding.
Extensive experiments show that SnapMLA achieves up to a 1.91× improvement in throughput on long-output decoding workloads while maintaining near-parity benchmark quality compared with the BF16 baseline on the evaluated reasoning and code-generation benchmarks.
These contributions position SnapMLA as an effective framework for efficient decoding of long-context MLA LLMs, ensuring both high computational throughput and the preservation of model accuracy.

\clearpage
\bibliography{neurips}
\bibliographystyle{plain}

\appendix
\clearpage

\section{Limitations}
\label{appendix:limitation}
SnapMLA is optimized for MLA decoding on NVIDIA Hopper-class GPUs, leveraging FP8 Tensor Cores, WGMMA, TMA, and Hopper-specific memory layouts. While these hardware-aware design principles are generally applicable, the reported kernel-level speedups are specific to this execution substrate and would require architecture-specific adaptations for earlier GPUs or alternative accelerators.

In addition, SnapMLA employs FP8 E4M3 quantization for the content portion of the MLA KV cache while preserving the RoPE component at higher precision. Our numerical analyses and benchmark results validate this mixed-precision strategy for the evaluated decoding workloads. Exploring more aggressive precision formats, alternative calibration policies, or substantially longer context lengths remains a promising direction for future investigation.

\section{Additional Related Work}
\label{appendix:additional_related_work}

\paragraph{FlashAttention-3 and FP8 Attention Acceleration}
The NVIDIA Hopper architecture introduces fourth-generation Tensor Cores, offering substantial throughput gains via FP8 computation. However, unlocking this potential requires adhering to strict hardware constraints: the FP8 WGMMA instructions mandate that the $\mathbf{V}$ matrix in the PV GEMM be contiguous along the sequence dimension ($k$-major). This requirement conflicts with the conventional head-contiguous memory layout used in standard attention implementations.

FlashAttention-3~\cite{shah2024flashattention3} enables efficient FP8 attention acceleration by resolving this layout mismatch without expensive offline data conversion. Instead of relying on hardware-specific memory instructions, it adopts an on-the-fly transposition strategy within the computation kernel. By dynamically reshaping the $\mathbf{V}$ blocks during the loading phase, it aligns the data layout with the Tensor Core requirements while pipelining memory access with computation to hide latency. Furthermore, to align the QK$^\top$ GEMM accumulator with the input requirements of the subsequent PV GEMM, FlashAttention-3 utilizes register-level permutations. By ensuring the operands match the specific layout required by WGMMA instructions, it achieves mathematical exactness while maximizing the utilization of FP8 compute units.

While FlashAttention-3 successfully accelerates standard MHA through this uniform FP8 pipeline, its design premises restrict its direct applicability to MLA. Specifically, MLA's decoupled RoPE component imposes heterogeneous precision requirements that are incompatible with a uniform FP8 data path. SnapMLA addresses this challenge by preserving the RoPE component in higher precision while quantizing the latent content path.

\paragraph{FlashMLA: Architecture-Aware MLA Decoding on NVIDIA Hopper GPUs}
FlashMLA~\cite{flashmla2025} is a specialized kernel engineered for the NVIDIA Hopper architecture to accelerate MLA decoding. Unlike traditional kernels that often suffer from memory-bound bottlenecks, FlashMLA is designed to maximize instruction throughput in compute-bound scenarios by strictly aligning algorithmic execution with hardware resources.

The technical core of FlashMLA lies in its asynchronous pipelining. By leveraging TMA and WGMMA instructions, the kernel enables direct matrix operations on shared memory, reducing pipeline bubbles caused by register movement and synchronization. It uses a warp-specialized scheduling model where data movement and computation are decoupled through parameterizable buffers, mitigating register pressure within the SM. Its fine-grained TMA-to-GEMM pipeline also allows GEMM operations to trigger immediately upon partial data arrival, helping maintain Tensor Core utilization. SnapMLA extends this line of work to FP8 MLA decoding by introducing RoPE-aware quantization, scale-aware PV reconstruction, and fused dataflow support.

\section{Preliminaries of Low-Bit Quantization}
\begin{figure}[h]
    \centering
    \includegraphics[width=1\linewidth]{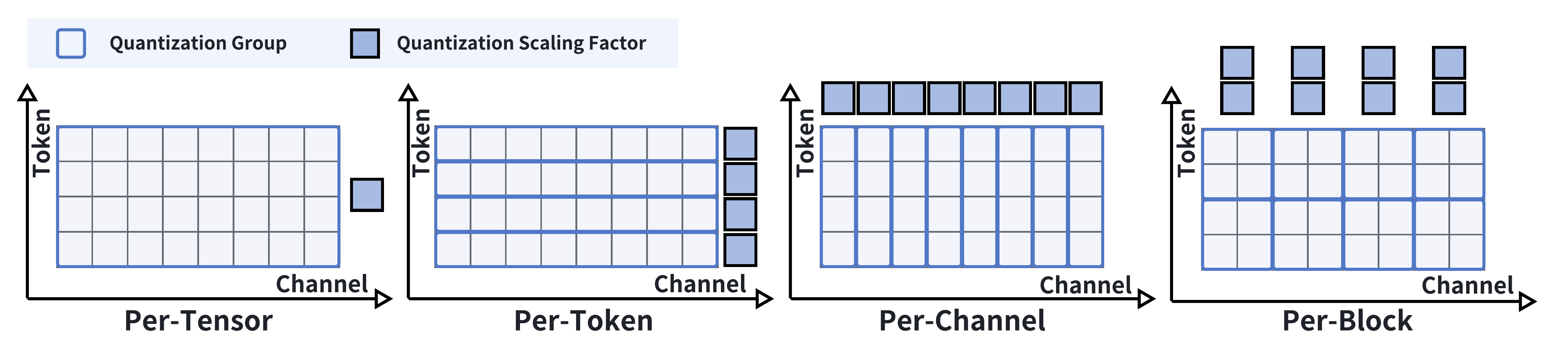}
    \caption{Illustration of various quantization granularities.}
    \label{fig:quant gran}
\end{figure}
Extensive prior research has focused on the quantization of LLMs, addressing both linear layers \cite{frantar2022gptq,lin2024awq,hooper2024kvquant,li2023fptq,xiao2025exploring,zhang2026locate} and KV caches \cite{li2024survey,hooper2024kvquant,liu2024kivi,su2025rotatekv,su2025akvq}. 
This technique not only reduces memory footprint and data transfer latency but also facilitates acceleration through low-precision computation kernels.
The evolution of hardware has driven a shift from integer (INT) to floating-point (FP) quantization \cite{shah2024flashattention3,micikevicius2022fp8,kuzmin2022fp8}. 
On older architectures like NVIDIA Ampere, INT quantization prevails, with accuracy enhanced by tailoring weight and activation distributions \cite{lin2024awq,ashkboos2024quarot}. 
Meanwhile, newer generations such as Hopper and Blackwell unlock native support for low-bit FP computation.
With its wider dynamic range, FP quantization inherently captures distribution characteristics more effectively \cite{van2023fp8}, yielding higher accuracy than INT at the same bit-width.

The granularity of quantization plays a significant role in determining accuracy, computational efficiency, and engineering feasibility. 
Typically, several levels of quantization granularity exist (as illustrated in Figure \ref{fig:quant gran}), each determined by the method used to apply scaling factors to the input tensor \( \mathbf{X} \in \mathbb{R}^{M \times N} \):
\begin{enumerate}[label=\textbf{(\arabic*)}, leftmargin=*]
    \item \textit{\textbf{Per-Tensor Quantization}}
    
    A single scalar scaling factor \(S \in \mathbb{R}^{1}\) is shared across the entire tensor:
    \begin{equation}
        \mathbf{X}_q = \text{round}\left( \frac{\mathbf{X}}{S} \right).
    \end{equation}
    This method minimizes metadata overhead but may compromise accuracy due to limited adaptability to the tensor's diverse values.

    \item \textit{\textbf{Per-Token Quantization}}
    
    An independent scaling factor is computed for each token (i.e., each row), resulting in a column vector \(S \in \mathbb{R}^{M \times 1}\):
    \begin{equation}
        \mathbf{X}_q^{(i,:)} = \text{round}\left( \frac{\mathbf{X}^{(i,:)}}{S_i} \right), \quad i = 1,\dots,M.
    \end{equation}
    Previous research has shown that significant outlier activation tokens \cite{su2025unveiling,sun2024massive} and KV cache outlier tokens \cite{su2025kvsink} exist in LLMs. 
    This approach captures cross-token activation variations, making it particularly effective for handling outliers.

    \item \textit{\textbf{Per-Channel Quantization}}
    
    In this scheme, each channel (i.e., each column) of the tensor is quantized independently using a separate scaling factor. 
    The scaling factors are organized as a vector \(S \in \mathbb{R}^{1 \times N}\):
    \begin{equation}
        \mathbf{X}_q^{(:,j)} = \text{round}\left( \frac{\mathbf{X}^{(:,j)}}{S_j} \right), \quad j = 1, \dots, N.
    \end{equation}
    This method allows for more flexibility in capturing variations across different channels \cite{hooper2024kvquant,liu2024kivi}, improving accuracy over per-tensor quantization.

    \item \textit{\textbf{Per-Block Quantization}}
    
    The tensor is partitioned into fixed \(B \times B\) sub-blocks (e.g., \(128 \times 128\) or \(64 \times 64\)), with each block assigned an independent scaling factor arranged in a matrix:
    \begin{equation}
        S \in \mathbb{R}^{\lceil \frac{M}{B} \rceil \times \lceil \frac{N}{B} \rceil}.
    \end{equation}
    The quantization is then applied per block:
    \begin{equation}
        \mathbf{X}_q^{(\text{block }(i,j))} = \text{round}\left( \frac{\mathbf{X}^{(\text{block }(i,j))}}{S_{i,j}} \right).
    \end{equation}
    This scheme balances accuracy with hardware-friendly block-wise computation, particularly beneficial in scenarios where parallel processing of smaller blocks is required.
\end{enumerate}

\label{appendix:low bit quant}

\section{Online Scale Fusion for Softmax Accumulation}
\label{Online Scale Fusion in the Softmax Function}

\paragraph{Scale Convention and Exactness}
Throughout this section, a quantized tensor \( X_q \) with scale \( \sigma_X \) represents the real-valued tensor \( X \approx \sigma_X X_q \), with scales broadcast along the corresponding GEMM reduction dimension. For QK computation, \( \sigma_q \) and \( \sigma_K \) correspond to the content scales \( S^{Q_c} \) and \( S^{K_c} \) used in the pre-scaled RoPE alignment. For PV computation in the absorbed MLA form, the value operand reuses the latent content cache; therefore, the per-token \( V \) scale \( S_V \) follows the same latent-cache scale convention as the content cache. We use \( \sigma_P^{(k)} \) exclusively for the dynamic quantization scale of the fused probability block at tile \( k \). The scale relocation \( P(S_V \odot V_q) = (P \odot S_V)V_q \) and the online state rescaling are algebraically exact up to finite-precision rounding; approximation is introduced by FP8 quantization of the content cache and fused probability blocks. In implementation, dynamic scales are lower-bounded by a small \( \epsilon \) before division to avoid zero-scale cases.

Block-wise \( P \) quantization introduces scale variations across blocks. To maintain arithmetic accuracy during tiled accumulation without incurring explicit dequantization overhead, we fold the probability-block quantization scale into the online Softmax accumulation updates. Specifically, we track two intermediate states during the tiled computation: the scaled normalization statistic \( L \), which tracks the accumulated sum of exponentials in the current probability-scale domain, and the partial attention output \( O \). The QK logits and the Softmax distribution are unchanged; scale handling is applied after exponentiation, when the per-token \( V \) scale is fused into the current probability block.

To ensure hardware efficiency and minimize computational overhead, we maintain \( O \) and \( L \) in the current probability-scale domain. Let \( \sigma_P^{(k)} \) denote the dynamic quantization scale of the fused probability block at tile \( k \), and let \( S_V^{(k)} \) denote the per-token scale of \( V \). The following update rules describe how the normalization statistic and the attention output are computed in this hardware-optimized pipeline:

\paragraph{Hardware-Optimized Online Fusion (With \( L \) and \( O \) Updates)}
For each block \( k \), let \( s_j \) be the restored QK logit and \( m^{(k)} = \max(m^{(k-1)}, \max_{j \in \text{Block } k} s_j) \). We first compute the unnormalized Softmax term \( e_j = e^{s_j - m^{(k)}} \), then form the fused probability term \( \tilde{e}_j = e_j S_{V,j}^{(k)} \) before block-wise quantization. Here, \( V_{q,j} \) denotes the quantized \( V \) value used by the FP8 PV GEMM. The states \( L^{(k)} \) and \( O^{(k)} \) are updated iteratively as follows:
\begin{equation}
    L^{(k)} = L^{(k-1)} \cdot e^{(m^{(k-1)} - m^{(k)})} \cdot \frac{\sigma_P^{(k-1)}}{\sigma_P^{(k)}} + \frac{\sum_{j \in \text{Block } k} e_j}{\sigma_P^{(k)}}
\end{equation}
\begin{equation}
    O^{(k)} = O^{(k-1)} \cdot e^{(m^{(k-1)} - m^{(k)})} \cdot \frac{\sigma_P^{(k-1)}}{\sigma_P^{(k)}} + \frac{\sum_{j \in \text{Block } k} \tilde{e}_j V_{q,j}}{\sigma_P^{(k)}}
\end{equation}
This formulation ensures that the output \( O \) and normalization statistic \( L \) are dynamically re-scaled to account for both the shifting maximum \( m^{(k)} \) and the block-specific probability scale \( \sigma_P^{(k)} \), thus eliminating the need for a separate dequantization pass.

\section{Supplementary Implementation Details for Quantized PV Computation}
\label{Supplementary Implementation Details for Quantized PV Computation}
\newcommand{\WGZERO}[1]{\textnormal{\texttt{WG0:}}~#1}
\newcommand{\WGONE}[1]{\textnormal{\texttt{WG1:}}~#1}
\newcommand{\Transpose}[1]{#1^\top}
\newcommand{\Quant}[2]{\mathrm{Quant}(#1;\, #2)}
\newcommand{\Max}{\mathrm{max}}
\newcommand{\Sum}{\mathrm{sum}}
\newcommand{\Exp}{\exp}
\newcommand{\Log}{\log}
\newcommand{\BlockReduceSum}{\mathrm{BlockReduceSum}}
\begin{algorithm*}[h]
\caption{Decoding Process with the Dual-WG Pipeline}
\label{alg:flashmla_fp8_dual_wg}
\begin{algorithmic}[1]
\scriptsize 

\Require $\mathbf{q} = [\mathbf{q}_c, \mathbf{q}_r] \in \mathbb{R}^{d}$ in HBM, with $\mathbf{q}_c \in \mathbb{R}^{d_c}$ (FP8) and $\mathbf{q}_r \in \mathbb{R}^{d_r}$
\Require Cached $\mathbf{K} = [\mathbf{K}_c, \mathbf{K}_r] \in \mathbb{R}^{N \times d}$ in HBM, with $\mathbf{K}_c \in \mathbb{R}^{N \times d_c}$ (FP8) and $\mathbf{K}_r \in \mathbb{R}^{N \times d_r}$; in absorbed MLA, the PV value operand reuses the latent content cache, denoted $\mathbf{V}_q \equiv \mathbf{K}_c$
\Require Fp8 scaling factors $\boldsymbol{\sigma}_{\mathbf{q}} \in \mathbb{R}$ (for $\mathbf{q}_c$) and $\boldsymbol{\sigma}_{\mathbf{K}} \in \mathbb{R}^{N \times 1}$ (for $\mathbf{K}_c$) in HBM
\Require $\boldsymbol{\sigma}_{\mathbf{q}}$ and $\boldsymbol{\sigma}_{\mathbf{K}}$ are pre-injected into $\mathbf{q}_r$ and $\mathbf{K}_r$
\Require Key block size $B_c=64$ with $T_c = \lceil \frac{N}{B_c} \rceil$, two warp groups: WG0, WG1.

\State Initialize $p_0, p_1 \in \mathbb{R}^{B_c}$ , $\gamma_0, \gamma_1$ , $m = -\infty$, $\sigma_p = 1.0$ in SRAM
\State Initialize $\mathbf{o}^L = \mathbf{o}^R = (0) \in \mathbb{R}^{\frac{d_c}{2}}$, $\ell = 0$ in Register

\For{$j = 0$ \textbf{to} $Tc - 1$ \textbf{step} 2}

    \State {Load $\mathbf{K}_0 = \mathbf{K}_j$ , $\mathbf{K}_1 = \mathbf{K}_{j+1}$ in SRAM}
    \State 
    \begin{minipage}[t]{0.48\linewidth}
        \textbf{\underline{Warp Group 0 (WG0)}}
        \begin{enumerate}[label=\arabic*:, leftmargin=1.5em, itemsep=0pt, parsep=0pt, topsep=2pt]
            \item $\mathbf{s}_j = \mathbf{q} \mathbf{K}_0^\top$
            \item $\mathbf{V}_0^L = \Transpose(\mathbf{K}_0^L)$ , $\mathbf{V}_0^R = \Transpose(\mathbf{K}_0^R)$
            \item $\mathbf{s}_j = \mathbf{s}_j \odot (\boldsymbol{\sigma}_{\mathbf{q}} \boldsymbol{\sigma}_{\mathbf{K}_0}^\top)$
            \item $m^{cur} = \Max(\mathbf{s}_j)$ , $m^{new} = \Max(m, m^{cur})$
            \item $p_j = \Exp(s_j - m^{new})$ , $\ell^{cur} = \Sum(p_j)$
            \item $p_j = p_j \odot \boldsymbol{\sigma}_\mathbf{K_0}$ , $m^{cur} = \Max(p_j)$
            \item $\sigma_p^{cur} = m^{cur} / 448.0$ , $p_j^{'} = \Quant(p_j, \sigma_p^{cur})$
            \item $\ell^{cur} = \ell^{cur} / \sigma_p^{cur}$ 
            \item $\gamma = \Exp(m - m^{new}) \sigma_p / \sigma_p^{cur}$
            \item $\mathbf{o}^L = \gamma \mathbf{o}^L$ , $\ell = \gamma \ell + \ell^{cur}$
            \item Store $\gamma_0 = \gamma$ , $m = m^{new}$ , $\sigma_p = \sigma_p^{cur}$
            \item \textbf{Arrive}($\gamma_0$Ready)
            \item Store $p_0 = p_j^{'}$
            \item \textbf{Arrive}($p_0$Ready)
            \item $\mathbf{o}^L = \mathbf{o}^L + p_j^{'} \mathbf{V}_0^L$
            \item \textbf{Wait}($p_0 \mathbf{V}_0^R$Issued)
            \item $\mathbf{o}^L = \gamma_1 \mathbf{o}^L$ , $\ell = \gamma_1 \ell$
            \item $\mathbf{o}^L = \mathbf{o}^L + p_1 \mathbf{V}_1^L$
        \end{enumerate}
    \end{minipage}%
    \hfill
    \begin{minipage}[t]{0.48\linewidth}
        \textbf{\underline{Warp Group 1 (WG1)}}
        \begin{enumerate}[label=\arabic*:, leftmargin=1.5em, itemsep=0pt, parsep=0pt, topsep=2pt]
            \item $\mathbf{s}_{j+1} = \mathbf{q} \mathbf{K}_1^\top$
            \item $\mathbf{V}_1^L = \Transpose(\mathbf{K}_1^L)$ , $\mathbf{V}_1^R = \Transpose(\mathbf{K}_1^R)$
            \item \textbf{Wait}($\gamma_0$Ready)
            \item $\mathbf{s}_{j+1} = \mathbf{s}_{j+1} \odot (\boldsymbol{\sigma}_{\mathbf{q}} \boldsymbol{\sigma}_{\mathbf{K}_1}^\top)$
            \item $m^{cur} = \Max(\mathbf{s}_{j+1})$ , $m^{new} = \Max(m, m^{cur})$
            \item $p_{j+1} = \Exp(s_{j+1} - m^{new})$, $\ell^{cur} = \Sum(p_{j+1})$
            \item $p_{j+1} = p_{j+1} \odot \boldsymbol{\sigma}_\mathbf{K_1}$, $m^{cur} = \Max(p_{j+1})$
            \item $\sigma_p^{cur} = m^{cur} / 448.0$,\\$p_{j+1}^{'} = \Quant(p_{j+1}, \sigma_p^{cur})$
            \item $\ell^{cur} = \ell^{cur} / \sigma_p^{cur}$ 
            \item $\gamma = \Exp(m - m^{new}) \sigma_p / \sigma_p^{cur}$
            \item $\mathbf{o}^R = \gamma_0 \mathbf{o}^R$ , $\ell = \gamma \gamma_0 \ell + \ell^{cur}$
            \item Store $\gamma_1 = \gamma$ , $m = m^{new}$ , $\sigma_p = \sigma_p^{cur}$
            \item Store $p_1 = p_{j+1}^{'}$
            \item \textbf{Wait}($p_0$Ready)
            \item $\mathbf{o}^R = \mathbf{o}^R + p_0 \mathbf{V}_0^R$
            \item \textbf{Arrive}($p_0 \mathbf{V}_0^R$Issued)
            \item $\mathbf{o}^R = \gamma \mathbf{o}^R$
            \item $\mathbf{o}^R = \mathbf{o}^R + p_{j+1}^{'} \mathbf{V}_1^R$
        \end{enumerate}
    \end{minipage}

\EndFor

\State {$\ell = BlockReduceSum(\ell)$}
\State {Merge $\mathbf{o}_L, \mathbf{o}_R$ and Normalize $\mathbf{o} = \mathbf{o} / \ell$}
\State {$\mathbf{L} = m  + \log(\sigma_p\ell)$}
\State {Write $\mathbf{o}$ , $\mathbf{L}$ to HBM}
\State {Return the output $\mathbf{o}$ and the logsumexp $\mathbf{L}$.}

\end{algorithmic}
\end{algorithm*}

\paragraph{The Scale Hazard in Double-Buffered Execution}
The original FlashMLA kernel utilizes a double-buffered pipeline, coordinated by two warp groups (WGs), to parallelize the \( \mathbf{P} \mathbf{V} \) computation. A key design feature is the inverted execution order of WG1: it computes the contribution from the second block (\( \mathbf{P}_1 \mathbf{V}_1^R \)) before the first block (\( \mathbf{P}_0 \mathbf{V}_0^R \)), maximizing instruction-level overlap and hiding memory latency.
However, this optimization introduces a fundamental numerical issue when using FP8 quantization. The problem arises from synchronizing the accumulator \( \mathbf{O}_{acc} \), which holds the running sum of partial results, with the dynamically scaled \( \mathbf{P} \) matrices.

\begin{itemize}
    \item \textbf{Problem 1 (Rescaling \( \mathbf{P}_0 \))}: After processing \( \mathbf{P}_1 \), \( \mathbf{O}_{acc} \) resides at the scale of \( \mathbf{P}_1 \). The original strategy rescales \( \mathbf{P}_0 \) to match this scale before accumulation. For FP8, this is problematic. The already quantized \( \mathbf{P}_0 \) is constrained within a severely limited dynamic range. Applying a large rescaling factor (when \( \sigma_{\mathbf{P}_1} \gg \sigma_{\mathbf{P}_0} \)) disrupts its value distribution, leading to irreversible precision loss.
    
    \item \textbf{Problem 2 (Rolling Back \( \mathbf{O}_{acc} \))}: The alternative—temporarily reverting \( \mathbf{O}_{acc} \) to the scale of \( \mathbf{P}_0 \) and restoring it later—is equally problematic. This approach requires bidirectional rescaling, involving multiplication by both the ratio \( \sigma_{\mathbf{P}_0}/\sigma_{\mathbf{P}_1} \) and its reciprocal \( \sigma_{\mathbf{P}_1}/\sigma_{\mathbf{P}_0} \). As a result, each scale becomes a denominator in one of these operations. When a significant disparity exists between \( \sigma_{\mathbf{P}_0} \) and \( \sigma_{\mathbf{P}_1} \), these ratios can explode or vanish, posing a severe risk to numerical stability.
\end{itemize}

\paragraph{Lossless Pipeline Reconstruction via Order Enforcement}
To resolve this issue, we abandon the inverted execution logic and enforce a strictly monotonic execution order for WG1: \( \mathbf{P}_0 \mathbf{V}_0^R \rightarrow \mathbf{P}_1 \mathbf{V}_1^R \).

This reconstruction guarantees that the accumulator’s scale updates follow a strict, unidirectional progression, aligned with the computation order. The bidirectional rescaling dependency—and the associated risk of division by extreme scale differences—is completely eliminated.

Although this serialization alters the original memory-compute overlap pattern, we recover the lost performance by fine-tuning inter-group synchronization barriers and instruction scheduling. The redesigned pipeline, detailed in Algorithm~\ref{alg:flashmla_fp8_dual_wg}, successfully balances the strict numerical stability required for FP8 computation with the high-throughput demands of a production-grade kernel.


\section{Benchmarks Details}
\label{appendix_benchmark}
The evaluation is conducted across multiple domains using the following benchmarks to ensure a thorough and comprehensive assessment:
\begin{itemize}
    \item \textbf{General QA:}  
    MMLU-Pro \cite{wang2024mmluprorobustchallengingmultitask}, a robust re-evaluated version of MMLU \cite{hendrycks2021measuringmassivemultitasklanguage} that corrects errors and reduces contamination.  
    MMLU-Redux \cite{gema2025mmluredux}, another high-quality variant of the MMLU benchmark.

    \item \textbf{Alignment:}  
    IFEval \cite{zhou2023ifeval}, an instruction-following benchmark consisting of a set of prompts with programmatically verifiable constraints, offering an objective score on the model’s fidelity to complex instructions.  
    Arena-Hard \cite{arenahard2024}, a benchmark from the Chatbot Arena platform for assessing a model’s helpfulness and conversational quality on difficult, open-ended user queries.  

    \item \textbf{Mathematical Reasoning:}  
    Olympiad-level mathematical benchmarks, including MATH-500 \cite{math500}, HMMT-25 \cite{HMMT25} (Harvard-MIT Mathematics Tournament), AIME-24 \cite{AIME24} and AIME-25 \cite{AIME25} (American Invitational Mathematics Examinations), and BeyondAIME \cite{bytedanceseed2025beyondaime}.
    
    \item \textbf{General Reasoning:}  
    GPQA-Diamond \cite{rein2024gpqa}, a benchmark for evaluating deep reasoning on graduate-level questions across several science domains.  
    ZebraLogic \cite{lin2025zebralogic}, consisting of classic logic grid puzzles that require multi-step deductive reasoning and constraint satisfaction.

    \item \textbf{Coding:}  
    LiveCodeBench (LCB) \cite{jain2025livecodebench}, a dynamic benchmark for evaluating coding problems, with specific problems ranging from 2408 to 2505.
\end{itemize}

\paragraph{Generated Length Statistics}
SnapMLA optimizes the autoregressive decoding stage, where reasoning and code-generation tasks can produce long outputs.
Table \ref{tab:runtime_length} reports the average generated lengths observed on DeepSeek-V3.1.
These statistics characterize the output lengths of the evaluated decoding workloads and show that several benchmarks exercise the attention kernel over thousands to tens of thousands of generated tokens.
They are intended to characterize long-output decoding behavior, rather than to replace dedicated long-context input-understanding benchmarks.
After FP8 quantization, the generated lengths remain close to their BF16 counterparts across benchmarks, with no consistent shortening trend.
This provides additional evidence that the small benchmark accuracy differences in Table \ref{tab:benchmarks} are not caused by shortened or truncated outputs.

\begin{table*}[t]
\centering
\caption{Average generated lengths of DeepSeek-V3.1 decoding workloads. These statistics characterize the long-output behavior of the evaluated benchmarks.}
\label{tab:runtime_length}
\resizebox{0.65\textwidth}{!}{%
\begin{tabular}{@{}lrrr@{}}
\toprule
Benchmark & BF16 & FP8 & Relative Diff. \\
\midrule
\multicolumn{4}{c}{\cellcolor[HTML]{EFEFEF}\textit{General}} \\
MMLU-Pro$_{\text{(acc)}}$ & 2,447 & 2,471 & +1.0\% \\
MMLU-Redux$_{\text{(acc)}}$ & 562 & 558 & -0.7\% \\
\midrule
\multicolumn{4}{c}{\cellcolor[HTML]{EFEFEF}\textit{Instruction}} \\
IFEval$_{\text{(strict prompt)}}$ & 680 & 672 & -1.2\% \\
Arena-Hard$_{\text{(hard prompt gemini)}}$ & 3,275 & 3,254 & -0.6\% \\
\midrule
\multicolumn{4}{c}{\cellcolor[HTML]{EFEFEF}\textit{Math}} \\
MATH-500$_{\text{(Mean@1)}}$ & 2,346 & 2,397 & +2.2\% \\
HMMT-25$_{\text{(Mean@32)}}$ & 16,618 & 16,989 & +2.2\% \\
AIME-24$_{\text{(Mean@32)}}$ & 11,909 & 11,616 & -2.5\% \\
AIME-25$_{\text{(Mean@32)}}$ & 15,208 & 15,327 & +0.8\% \\
\midrule
\multicolumn{4}{c}{\cellcolor[HTML]{EFEFEF}\textit{Reasoning}} \\
GPQA-Diamond$_{\text{(Mean@16)}}$ & 9,183 & 8,942 & -2.6\% \\
ZebraLogic$_{\text{(Mean@1)}}$ & 5,091 & 4,972 & -2.3\% \\
\midrule
\multicolumn{4}{c}{\cellcolor[HTML]{EFEFEF}\textit{Coding}} \\
LCB (24.08-25.05)$_{\text{(Mean@4)}}$ & 13,034 & 13,047 & +0.1\% \\
OJBench$_{\text{(Mean@1)}}$ & 21,174 & 22,041 & +4.1\% \\
\bottomrule
\end{tabular}
}
\end{table*}

\section{Experiments on Numerical Accuracy}
\label{appendix:numerical_accuracy}
\paragraph{Settings}
To further evaluate the quantization accuracy of SnapMLA, we performed a layer-wise numerical analysis using real inference data from \textit{LongCat-Flash-Thinking}. 
We compare the attention outputs from SnapMLA with the BF16 ground truth, evaluating the discrepancies using three metrics: Root Mean Square Error (RMSE), Cosine Difference ($1 - \text{cosine similarity}$), and Relative L2 Error, across all layers.
To analyze the impact of quantization errors under different granularities and provide a comprehensive evaluation of SnapMLA's effectiveness in reducing errors, we also set up four alternative quantization configurations, as detailed in Table \ref{tab:Numerical Accuracy}.

\begin{table}[t]
    \centering
\caption{Different quantization configurations for KV cache in the numerical accuracy analysis.}
\label{tab:Numerical Accuracy}
\resizebox{\textwidth}{!}{%
\begin{tabular}{@{}c|l|cc@{}}
\toprule
\multirow{2}{*}{\textbf{Method}} & \multicolumn{1}{c|}{\multirow{2}{*}{\textbf{Configuration}}} & \multicolumn{2}{c}{\textbf{KV Cache Quantization}} \\ \cmidrule(l){3-4} 
 & \multicolumn{1}{c|}{} & \textbf{Content Part} & \textbf{RoPE Part} \\ \midrule
\textit{\textbf{SnapMLA (Ours)}} & \textit{\textbf{Per-Token RoPE-Aware}} & \textit{\textbf{Per-Token}} & \textit{\textbf{w/o quantization}} \\
Config A & Per-Token RoPE-Unaware & Per-Token & Per-Token \\
Config B & Per-Tensor Static RoPE-Aware & Per-Tensor Static (Fixed Scale 1.0) & w/o quantization \\
Config C & Per-Tensor Dynamic RoPE-Aware & Per-Tensor Dynamic & w/o quantization \\
Config D & Per-Block RoPE-Aware & Per-Block & w/o quantization \\ \bottomrule
\end{tabular}%
}
\end{table}
\begin{figure*}[t]
    \centering
    \includegraphics[width=\textwidth]{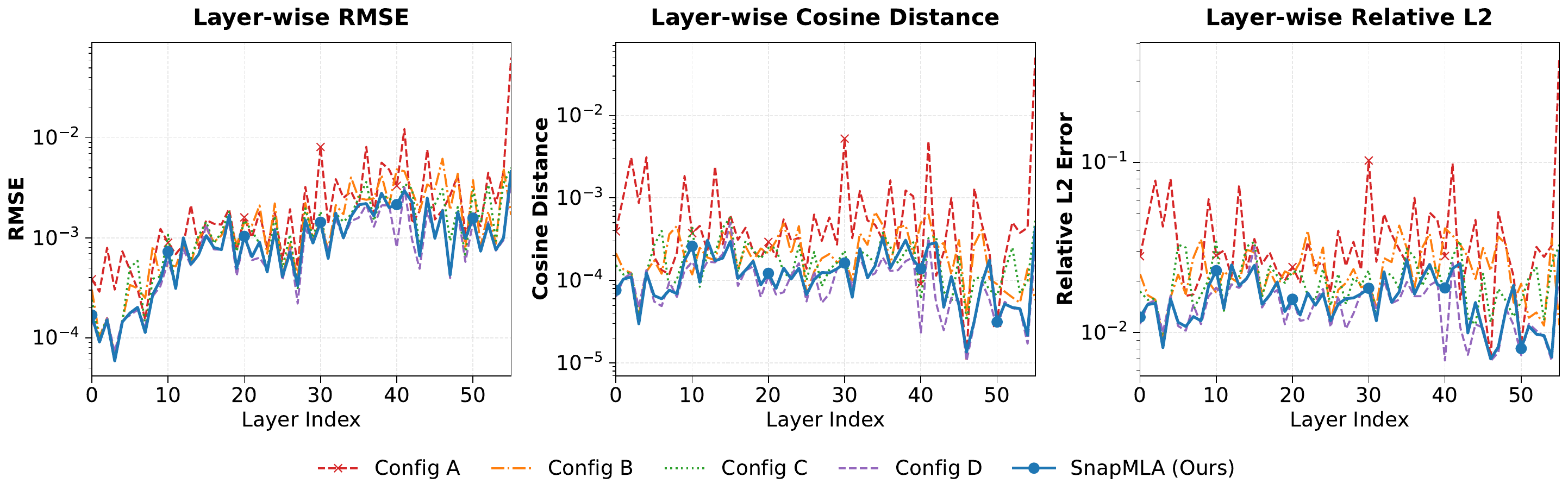}
    \caption{Layer-wise numerical fidelity analysis (context length = 32k). 
    For details on the quantization configurations, please refer to Table \ref{tab:Numerical Accuracy}.
}
    \label{fig:layerwise_error}
\end{figure*}
\paragraph{Analysis}
The layer-wise error comparison, shown in Figure \ref{fig:layerwise_error}, reveals two key insights.
First, including RoPE term in the quantization process \textit{(Config A)} results in a substantial increase in error metrics, particularly in the deeper layers. 
This "error explosion" empirically demonstrates the sensitivity of positional embeddings, underscoring the necessity of our RoPE-aware quantization.
Second, as highlighted in the zoomed-in insets, coarse-grained quantization strategies \textit{(Config B\&C)} struggle to capture the dynamic range variations across tokens. 
Even the block-wise strategy \textit{(Config D)} exhibits slightly higher degradation compared to our approach. 
In contrast, our method, which utilizes RoPE-Aware Per-token quantization, consistently delivers the lowest error rates across all metrics, maintaining numerical accuracy on par with the BF16 baseline.

\begin{figure*}[t]
    \centering
    \includegraphics[width=1\textwidth]{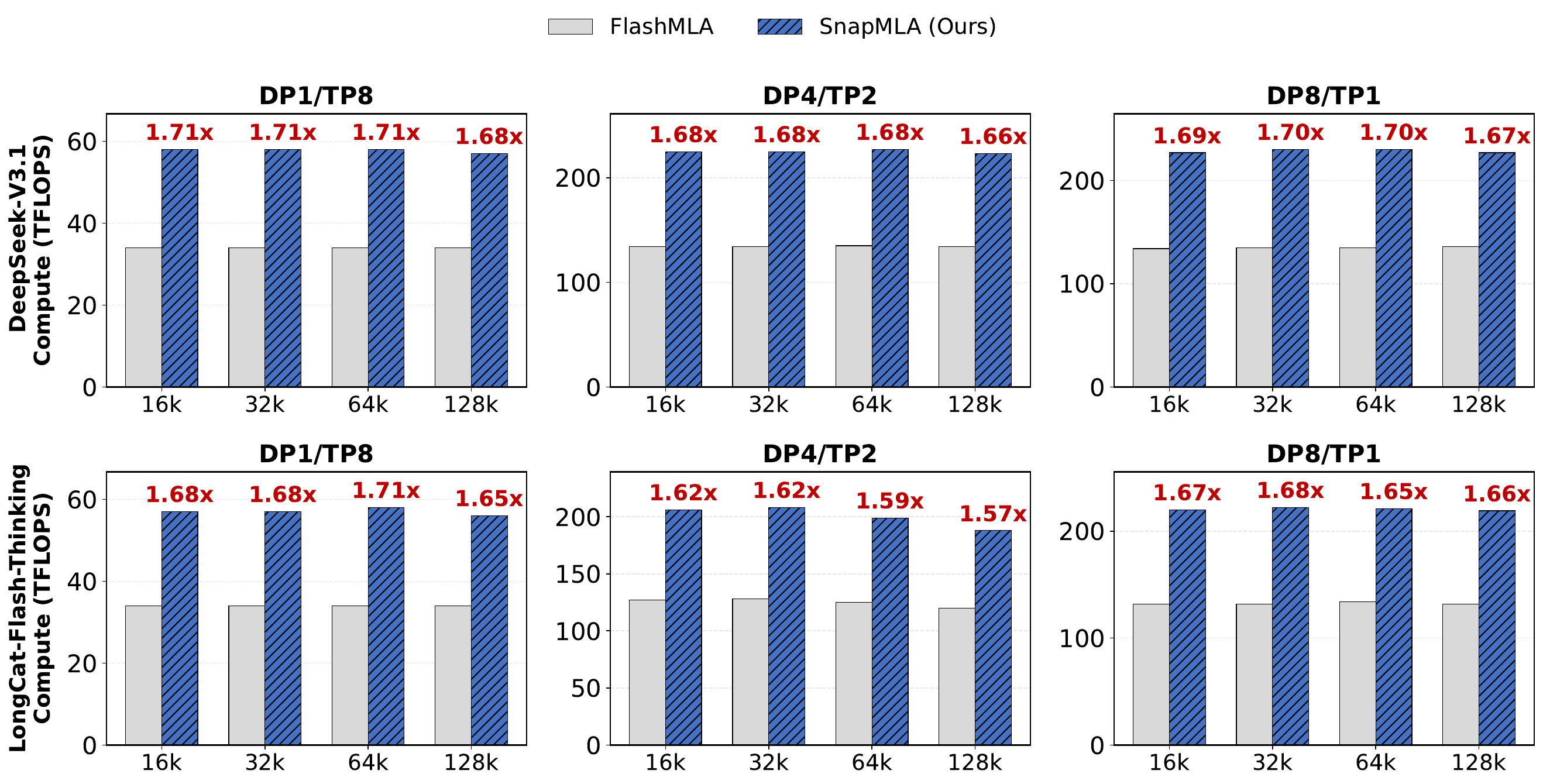} 
    \caption{Kernel-level compute performance (TFLOPS). We measure the compute throughput of SnapMLA (blue hatched) versus the FlashMLA baseline (gray) across varying sequence lengths. The workload configurations are derived from the corresponding end-to-end DP/TP settings. Our kernel closely tracks the trajectory of the effective theoretical peak.}
    \label{fig:kernel_tflops}
\end{figure*}
\section{Kernel Efficiency and Roofline Analysis}
\label{appendix:roofline_analysis}
To gain a deeper understanding of the sources of the efficiency gains, we perform a comprehensive analysis of the compute throughput of the SnapMLA kernel under end-to-end workloads. 
We now derive the effective theoretical peak performance.
The computational core of the MLA kernel consists of sixteen FP8 tiles (content term) and one BF16 tile (RoPE term), with the equivalent computational cost in BF16 units reduced from 17 to \(16/2 + 1 = 9\).
Given the theoretical BF16 peak of 148 TFLOPS for the GPU we used, the effective FP8 peak performance is calculated as:
\begin{equation}
    \text{Peak}_{\text{effective}} = 148 \times \frac{17}{9} \approx 279.6 \text{ TFLOPS}
\end{equation}
As shown in Figure \ref{fig:kernel_tflops}, our kernel performance closely aligns with this effective peak, demonstrating that the overhead from pipeline reconstruction and layout transformation is negligible.

\begin{figure*}[t]
    \centering
    \includegraphics[width=\textwidth]{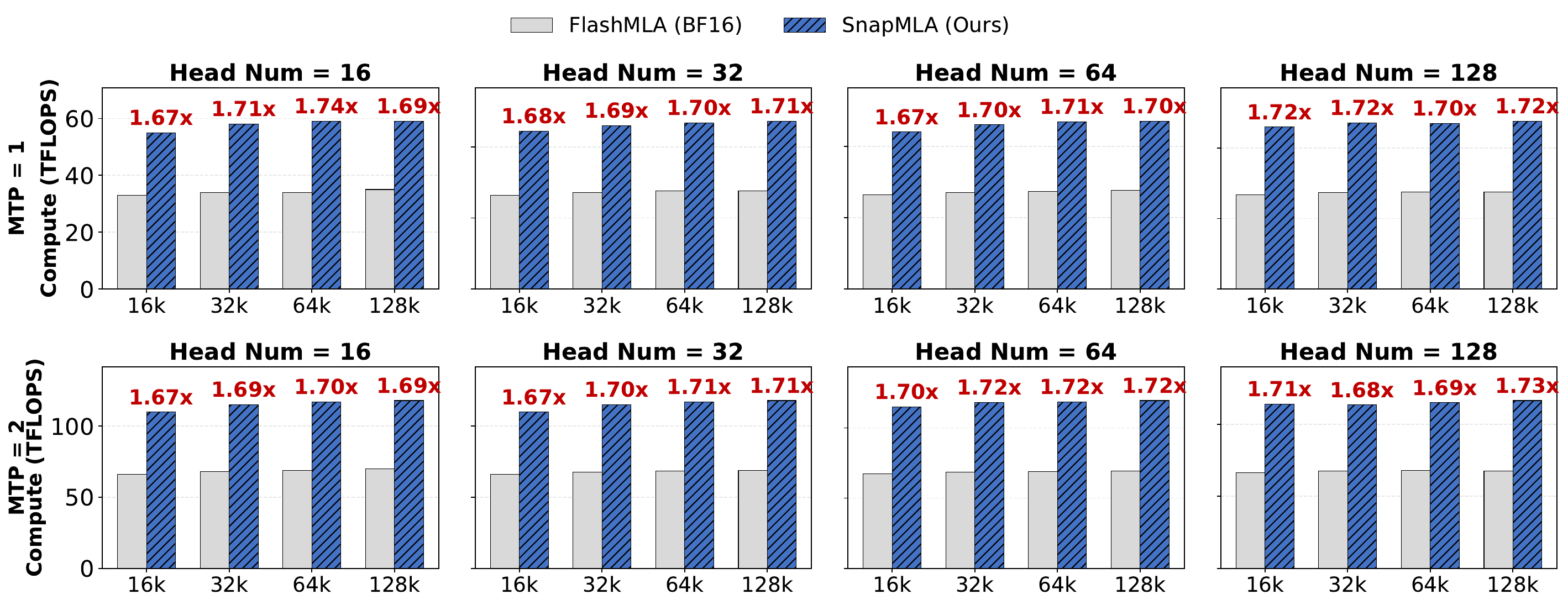} 
    \caption{Kernel performance across different input configurations. 
    We evaluate the compute throughput (TFLOPS) with a fixed batch size of 32, varying the number of heads from 16 to 128 and the query token length (MTP, rows) from 1 to 2. 
    The results show that kernel performance improves as the number of heads increases, stabilizing at higher configurations. Additionally, increasing the query length (MTP=2) yields a slight throughput gain compared to the standard decoding setting (MTP=1). 
    Across these configurations, SnapMLA consistently outperforms the baseline.}
    \label{fig:kernel_configurations}
\end{figure*}
\section{Sensitivity to Input Configurations}
\label{appendix:sensitivity_input}
In this section, we assess the kernel's robustness by evaluating its performance across various input configurations, including the number of heads (\( H \in \{16, 32, 64, 128\} \)) and multi-token prediction (MTP) settings (\( MTP \in \{1, 2\} \)), with a fixed batch size of 32. 

As illustrated in Figure \ref{fig:kernel_configurations}, throughput improves as the number of heads increases, reaching saturation at \( H \ge 64 \), where it attains approximately \textit{\textbf{85\% of the effective theoretical peak}}. Moreover, increasing the query length (MTP=2) provides a moderate performance boost. 
Throughout all configurations tested, and particularly for standard settings (\( H \ge 32 \)), SnapMLA consistently outperforms the baseline, demonstrating its robustness across varying input conditions.
\section{Evidence for Design Choices}
\label{appendix:design_evidence}

SnapMLA is designed and evaluated as an integrated decoding pipeline rather than as a collection of independently removable optimizations. Several of its components are coupled by the MLA shared latent KV representation and the Hopper FP8 WGMMA layout constraints. As a result, removing one component can produce an invalid kernel, a substantially different data path, or an unfair comparison rather than a clean ablation point. We therefore interpret the current evaluation as evidence for the integrated co-design, not as a complete factorial ablation of every kernel subcomponent.

\paragraph{RoPE-Aware Per-Token Quantization}
Appendix \ref{appendix:numerical_accuracy} provides the most direct evidence for this design choice. The layer-wise numerical analysis compares SnapMLA with alternative KV-cache quantization configurations, including RoPE-unaware quantization and coarser content quantization granularities. The results show that quantizing the RoPE component or using coarser content granularity leads to larger attention-output errors, supporting the choice of retaining the RoPE part in higher precision while applying per-token FP8 quantization to the content part.

\paragraph{Quantized PV Pipeline Reconstruction}
The PV reconstruction addresses the scale-layout mismatch caused by applying per-token scales to the MLA latent cache under the k-major layout required by FP8 WGMMA. The kernel-level analysis in Appendix \ref{appendix:roofline_analysis} shows that the resulting FP8 MLA kernel tracks the effective hardware roofline under end-to-end workload configurations. This suggests that the additional scale handling, probability-block quantization, and layout transformation needed by the reconstructed pipeline do not dominate runtime in the evaluated settings.

\paragraph{End-to-End Dataflow Optimization}
The end-to-end throughput results in Figure \ref{fig:throughput} reflect the combined effect of FP8 KV-cache storage, quantized PV computation, and fused data movement under realistic DP/TP serving configurations. The input-sensitivity study in Appendix \ref{appendix:sensitivity_input} further shows that the kernel maintains its advantage across relevant head-count and query-length configurations. Together, these results support the practical effectiveness of treating quantization, computation, and data movement as a coupled serving pipeline.


\end{document}